# Laser Ranging Based Intelligent System for Unknown Environment Mapping

T.H.M.N.C. Thelasingha[1], U.V.B.L. Udugama[2], E.M.S.P. Ekanayake[3],
G.M.R.I. Godaliyadda[4], M.P.B. Ekanayake[5], B.G.L.T. Samaranayake[6],
J.V. Wijayakulasooriya[7]

**Abstract:** Autonomous systems are used in many applications like unknown terrain mapping, Reconnaissance, Explorations etc. in areas of impossible human access. They require an accurate and practically realizable Intelligent Navigation System (INS), which can handle the complexity in the environment and necessarily be ready for mobile platform implementation. Also, a method of error free localization is a major requirement for the practical realization of such a system. This work outlines implementation of a system, simple enough to be implemented on a mobile platform, focused on provision of a logical and computationally efficient algorithm which uses measurements from a Laser Range Finder (LRF) as input. Usage of LRFs for mapping is a trending research area in contrast to the usage of Ultrasonic Sensors with errors like cross talk and specular reflections or vision systems that are power hungry. As, the LRF readings too contain some noise due to reflectivity problems, methods were developed for pre-processing the measurements. INS analyses the data spectrum intelligently to determine the obstacle free path that should be traversed, which will in turn enable the mobile platform to canvas the whole environment and construct the map. Also, the localization methods have been implemented through usage of similarity transform and particle filter. The conceptual design was simulated in Python™ on Windows™ and was tested for robustness in artificially generated environments. Then the system was emulated in Python™ with real environment data fed from the LRF in real-time. Then it was implemented in the embedded system level in a Raspberry pi3™ on a 3WD Omni-directional mobile platform, and tested for real environments. Also, the separately running localization algorithm has been used to correct the generated map and the position of the platform. The system was able to generate a complete two-dimensional map of the locale accurately. A quantitative justification of the proposed methodology is presented in a comparative analysis of the execution time.

## 1. Introduction

Environment mapping has been a key fact in endeavours like space explorations, excavations, disaster relief and reconnaissance applications. For that automation is highly essential when such tasks are carried out areas with restricted human access. Mapping an unknown environment autonomously has been an engineering problem that many researches have been focussed[1] [2] on for years. For it to be successful, development has to be done in many fronts.

Initially a correct sensing method has to be implemented to sense the geometry of the environment. Many of the published work include the use of RADARs [3], SONARs [4], Vision systems and Laser Range Finders (LRFs) [5]. The use of methods like and SONARs are convenient for simple distance sensing but when considered for applications like mapping they contain much noise. Although vision sensors provide more accurate and vivid data spectrum of the environment, the computational power required will make them less feasible for mobile platform implementation. In this work, a low cost LRF

Mr. T.H.M.N.C. Thelasingha, B.Sc. Eng. Undergraduate at Dept. of Electrical & Electronic Engineering, university of Peradeniya, Sri Lanka
Mr. U.V.B.L. Udugama, B.Sc. Eng. Undergraduate at Dept. of Electrical & Electronic Engineering, University of Peradeniya, Sri Lanka
Ms. E.M.S.P. Ekanayake, B.Sc. Eng. Undergraduate at Dept. of Electrical & Electronic Engineering, University of Peradeniya, Sri Lanka
Eng.(Dr.) M.P.B. Ekanayake, B.Sc.Eng. (Peradeniya), Ph.D. (Texas Tech,US), AMIE(Sri Lanka), Senior Lecturer at Dept. of Electrical & Electronic Engineering, University of Peradeniya, Sri Lanka
Eng. (Dr.) G.M.R.I. Godaliyadda, B.Sc. Eng. (Peradeniya), Ph.D. (NUS,Sg), ,AMIE (Sri Lanka), Senior Lecturer at Dept. of Electrical & Electronic Engineering, University of Peradeniya, Sri Lanka
Eng.(Dr.). B. G. L. T. Samaranayake, BSc Eng. Peradeniya, Tech. Lic & PhD KTH(Sweden), Senior Member IEEE, AMIE(SL), Senior Lecturer at Dept. of Electrical & Electronic Engineering, University of Peradeniya, Sri Lanka
Eng. (Dr.) J.V. Wijayakulasooriya, B.Sc Eng, (Peradeniya), Ph.D. (Northumbria,UK), MIE (Sri Lanka), Senior Lecturer at Dept. of Electrical & Electronic Engineering, University of Peradeniya, Sri Lanka [6] has been used as the distance sensor. A high-end sensor like Microsoft Kinect™ will render much accurate data but to keep the implementation realizable a simple sensor has been selected.

Another major problem is the method of traversing the unknown area. Various approaches like mobile robots and quad-rotors [7] are possible solutions. But considering the accuracy and convenience in positioning and manoeuvrability an omni-directional 3WD mobile platform [8] has been used as the mapping agent in this work. It has been developed completely in house. And its size and the instant mobility in any direction due to omni directional nature has allowed it to traverse many complex environments effectively. A state feedback linearizing nonlinear controller has been implemented so that the platform can adjust itself to a given reference coordinate and orientation.

Additionally, a convenient sensor has been selected and a method of motion is developed. Most approaches lack a correct navigation system [1] where the agent can be guided through the obstacles in the environment so that it canvases the whole area. Obstacle detection [9] and traversable space identification methods [10] have already been developed. Also, algorithms for finding the minimum cost path has been developed for some time [11].

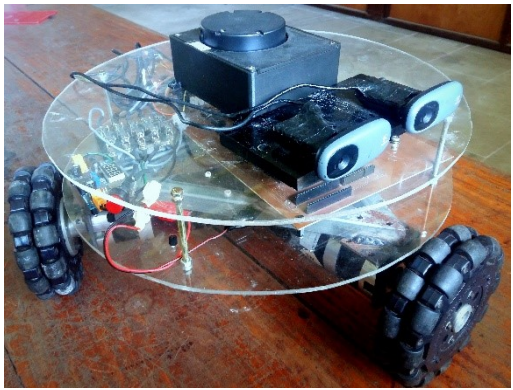

**Figure 1 – The Practical Realization**

But an approach to intelligently analyse the data from the distance measurements and decide the direction to be explored is a requirement. Hence a separate navigation algorithm has been developed which will analyse the information spectrum from the sensor and decides the path to be taken by the mobile platform to create a complete 2D overlay of the locale.

Here, to construct the map, the mobile platform traverses the area as guided by the navigation algorithm. And also the current position and orientation of it will be computed and fed to the navigation system in order to find the direction it should travel in the next moment. But the errors in the measurements and actuation are always possible. Also, the encoder based position is prone to errors due to wheel slipping and data transmission errors. Hence, for correct localization and to provide a correction for the map data generated from the LRF, a separately running simultaneous localization and mapping (SLAM) [12] algorithm based on Monte Carlo approach has also been developed. It uses the LRF sweeps from short time frames when the agent traverses in its trajectory and correct the encoder based position estimate in real time.

The combined implementation of the above three aspects namely, the hardware setup (omni directional mobile platform and the LRF), the navigation and path planning algorithm and the SLAM algorithm has allowed to develop a successful attempt in generating a 2D overlay of an unknown obstacle filled environment. For the task of extension of the 2D map to a complete 3D map, incorporating a vision system would be an effective approach. Further research in this work would include such a combination.

## 2. The Hardware Implementation

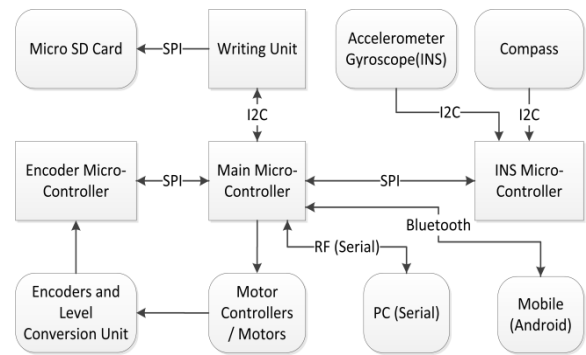

**Figure 2 – The complete embedded system architecture**

### 2.1 The Omni Directional Mobile Platform

The omni-directional mobile platform is the practical realization on which the proposed work has been implemented. It has been designed and constructed completely in house. A separate embedded system architecture has been developed, so that the control algorithms and also sensor fusion algorithms can be implemented on general purpose micro controllers. Also, the electronics of the system including the electronics of motor controlling and other auxiliary systems has been implemented on a single board with sensor modules.

Here, it is to be noted that the usage of GPS is avoided as in the scenarios like explorations and excavations which the equipment is developed for, would not guarantee the availability of GPS every time. An accurate estimation of the position and orientation of the mapping agent (pose or state) is obtained by fusing data from three different sensors that detect the orientation. The following sensors were used: a gyroscope (MPU6050), a compass (HMC5883L) and optical shaft encoders (OSE) (model - E50050030) attached to the wheels. To fuse the data from the above sensors, a centralized Kalman filter (CKF) was implemented in a general-purpose micro controller. As odometrical measurement system is coupled with an inertial measurement unit and a compass the accuracy of the position estimate has been improved much, while the system moves in different motion conditions like slipping and jerky motion. As controller strategy, a nonlinear state linearization controller has been implemented on the main micro controller. so that the ODR platform can be guided along any given trajectory while following a given orientation profile.

## 2.2 The Laser Range Finder (LRF) Data Processing

The LRF used, has a Neato laser sensor[6] driven by the Xv-11 LIDAR controller. The used LRF is a popular low-cost range finder compared to others in the market and has been used for many research and experimental applications. Although the accuracy is much higher than SONAR, or simple vision systems, it also has some deviations in the measured distances. The error characteristics of the LRF is shown in the Fig. 3.

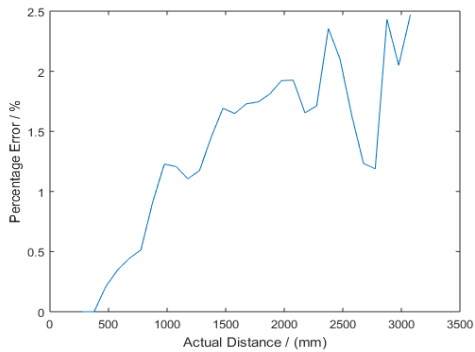

**Figure 3 – The Error Characteristic of LRF**

The laser sensor works by projecting a pulse stream of lasers around and measures the time it takes to reflect back. The distance to the reflection point can be accurately calculated (to 1 mm) from the time of flight the sensor then encodes the distance data into a packet and forward it to the LIDAR controller. The data packets are then decoded in the controller and relayed through the serial port up to the host device which is the embedded system that the Intelligent Navigation System (INS) implemented on. As the percentage error is small, in distances less than 3 m, the LRF measurement can be used as accurate estimation of distance. Separate method has been developed to mitigate the noise in the LRF measurements. Most of it was due to the reflectivity effects on the object surfaces and also the variation of the rotational speed of the LRF. To prevent overlapping of data points and incorrect angle measurements, A PID controller is implemented in the XV LIDAR controller to regulate the rotational speed of the sensor to the optimum speed of 200 rpm, where the device can send out a distance data point per a degree of rotation. Many of the reflectivity distortions could be rectified by sampling through multiple realizations. Here it was about five realizations, considering speed of operation and the power consumption.

A main capability that the system was intended to poses was its ability to be versatile and to be dynamically adaptive to any mobile platform. Hence, at the scanning step, the LRF is powered ON only at the scanning moment otherwise it is powered OFF, such that lesser power is consumed in the case of mobile implementation.

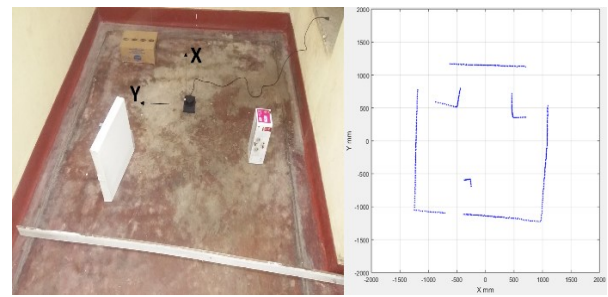

**Figure 4 – A Polar LRF Sweep of 360º**

## 3. The Intelligent Navigation System

The Intelligent Navigation System (INS) is the entity which analyses the data from the sensors and directs the mobile platform to the necessary place to explore [17]. The INS was

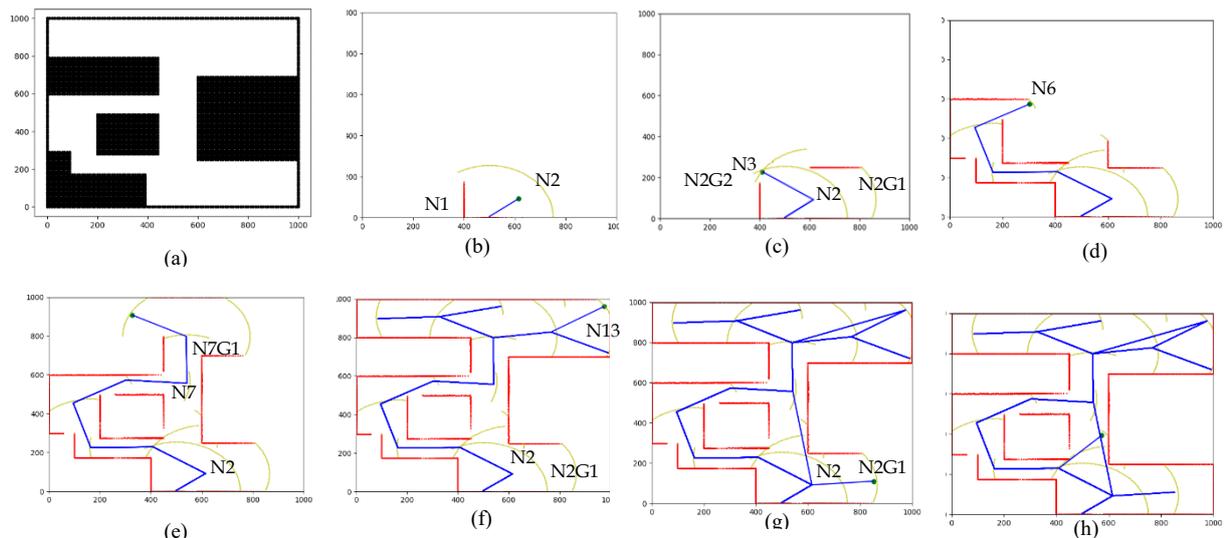

**Figure 5 – Simulation of Navigation Algorithm**

run for the simulation environment in Fig. 5(a) and the result was generated in the steps shown in the Fig. 5(b) to Fig. 5(h). It should be noted that the anchor points where the laser sweeps are done, are named as 'nodes' and respective nodes, which the agent can traverse to and from each other, are called as 'neighbours'. The procedure of the INS can be analysed through following explained steps,

1) Step 01: All the storage classes and data structures are initialized and current position and orientation is read. A node is created in the current position namely labelled as N1 in Fig. 5(b). (if this is not the first node, new node is also added as a neighbour for the previous node and vice versa) scanning is done for 360° initially because agent has to localize and decide which direction to explore. The list of distance data per degree is obtained from the LRF.

2) Step 02: Next the gaps and the walls should be correctly identified. Gap points are selected as data points with a relative distance from the agent, which is longer than a given length. Because those places can be assumed as places where it can be expected to find new information about the environment. The distance data is analysed to classify them as walls or gaps. At the end of each classification, its properties (width and centroid) are calculated. Here in the example, a large gap is detected from the scan at N1 hence the agent is directed towards the centroid of it. Hence the scanning is done again at the node N2, where 2 gaps have been detected. Here in Fig.5(c), it can be seen that the agent has been directed towards the minimum width gap first which is N2G2. This approach is justified in the latter part of the document. Here it should be noted that the scanning is done forward biased after the first scan. It is a 300° scan ±150° around the agents heading direction. In Fig. 5(d), it can be seen that the agent has now travelled up to the node N6 and further to node N7 in Fig. 5(e).

3) Step 03: It can be observed that direct travelling from N2 to N7 is possible. Hence, those two nodes should be neighbours. Also, it can be noticed that a scan down towards N2 from N7 is not necessary as the whole environment in between the nodes have been already mapped through the scans done at the respective nodes. Hence, to handle those kinds of situations a separate 'neighbour identification' is done as,

A.) If any two nodes are closer than twice the scan range, the whole area in between nodes can be assumed to be explored. Hence above situation occurs.

B.) Each gap of the corresponding two nodes are considered and checked if they intersect the line joining the two nodes. They are set as explored gaps as there is no new un-explored data in side those gaps, which are in between known nodes.

C.) Also if the widths of those gaps are larger than the size of the agent, they can be travelled through. Hence, the corresponding nodes are considered neighbours of each other (where node to node travelling is possible).

After the above analysis, the agent is now directed towards N7G1 because that is the information lacking area in the map. But INS has now learned that there is a path available to N2 from N7, without travelling or exploring any gap between them. This is kept track by constantly updating the travel cost map from each and every node to current node. The cost of travel (distance) is calculated to each node from the current node, through the neighbouring nodes. The minimum cost, unexplored node is selected and it is analysed to find the minimum width unexplored gap in its corresponding gaps. But the gap should be wide enough so that it can be traversed through. The minimum cost unexplored node obviously is the node where the agent is currently positioned. But if it does not contain any unexplored gaps, means a dead end. Then the INS intelligently decides to travel to the minimum cost node instead of just backtracking to find an unexplored gap. (This minimum cost node is the first approach and the narrowest gap first approach is justified in next section)

4) Step 04: In the next step, Fig. 5(f) after agent has explored the upper part of the map, at node N13, it has decided through the above explained process in step 03, to go back to unexplored node N2 to the gap N2G1. As of now the INS knows the path from N7 to N2, hence it does not unnecessarily backtrack, but travels through middle part of the environment to node N2 as observable from Fig. 5(g). The minimum cost path is found to the selected node employing the Dijkstra's Shortest path algorithm [11]. The reference point to travel is generated for the agent to follow the path intended. After node N2, it explores another unexplored gap at N3 and finishes up mapping as in Fig. 5(h)

The algorithm has been implemented so that it uses minimum amount of external complex libraries. Also, program loops have been avoided as much as possible and functions have been implemented so that it will be faster in execution. Conditional switches have been used to handle errors caused by data dependencies. The lesser processing power needed and the usage of simple functions and libraries have made it much more feasible to implement the system on a mobile platform with limited resources.

## 4. Improvement of Mapping Efficacy

When the efficacy of mapping is considered, two key factors are the accuracy of the map and the time it takes to map the whole area. The accuracy of the map highly depends on the accuracy of the LRF measurements and also correct estimation of current position relative to the locale. The control strategy deployed in the agent will determine the accurate positioning of the platform to the coordinates directed by the INS. Hence to correctly localize the agent a special SLAM algorithm has been employed. It uses the data from the LRF and compute necessary corrections needed for the position of the agent as well as the generated map.

Apart from the above approaches, the INS has been developed on several logics that improves the mapping time. one is the scanning method. Initially at the agents first scan of the environment, it is done for a full circle of 360°, so that the agent can localize itself and correctly decide that where should it explore first. But the scans followed after the initiation are done forward biased around ±150° around the heading direction. This approach limits the agent of collecting existing information again in the area it just explored and keep the information spectrum forward biased.

Another approach is the continuously updated cost of travel (distance) map of the nodes. If there are no gaps to explore in the current node the traditional approach is to backtrack and find the unexplored gaps in the past nodes according to the order of traversing. But in the proposed INS next node to explore is decided by analysing the cost of travel to the nodes and selecting the unexplored node with minimum cost of travel. This comes much effective in complex environments, when there exist two or more paths for node to node traversing, which are identified through the process of 'neighbour identification' explained in the above section. Hence, if the agent has found a lesser cost path to a past node, in its path planning, it will effectively utilize the found path instead of simple backtracking. It can be seen clearly in the example in Fig. 5, where a path has been found from N2 to N7 without an explicit scan.

Another approach towards improving mapping efficacy is selection of the minimum width gap to explore first. After selecting the respective node to travel as explained above, a gap need to be selected in the respective node.

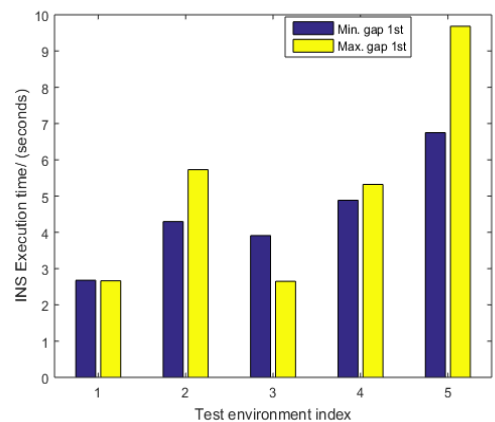

**Figure 6 – INS execution time comparison**

Here it's done under the general assumption that in the real world a smaller gap is much probable to lead to a convergence and reach a dead-end, unlike a wider gap. Hence, the minimum width gap is selected to explore before heading out in to the wider area where the probability of convergence is much lower. As a quantitative analysis of the efficacy, figure 6 shows a comparison of the execution times of the INS in 'Min. gap vs. Max. gap first' approaches for five test environments. From the comparative analysis, it can be concluded that the min. gap first approach is an efficient method of exploration.

## 5. Localization - SLAM

The accurate localization or else knowing the exact position is critical in the mapping task. As the existing GPS systems provide accuracy of few meters and also fails in indoors, SLAM approach has been utilized as a localization method [16]. Although the ussage of extended Kalman filter for SLAM is popular [13], due to its high computational complexity, solutions based on similarity transforms and particle filter (Monte Carlo Localization) [14] [15] based methods have been used in this work.

In both methods, encoder readings are used to estimate the position, then the LRF data is used to correct the position estimate. Finding the relevant features/objects using LRF is done using differentiation.

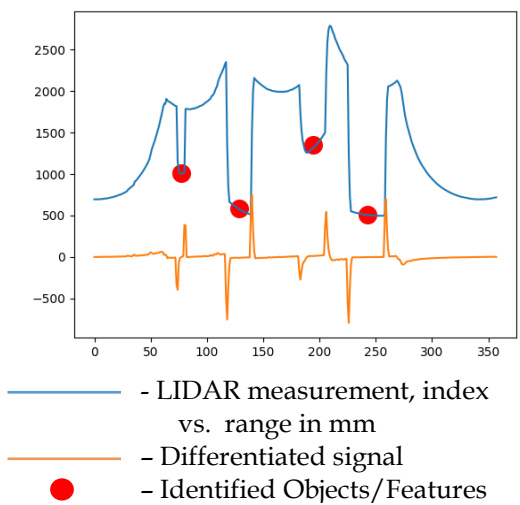

——— - LIDAR measurement, index vs. range in mm
——— – Differentiated signal
● – Identified Objects/Features

**Figure 7 – Feature Identification**

### 5.1 Similarity Transform

This method is based on finding an optimized parameter estimation for the transformation, to transform the agent back to its correct position. It has been assumed that the erroneous localization is occurred when it is traveling on curved paths. The detected feature positions can be used for correction. Correspondences between erroneous object positions and previously determined positions of the same object are found through similarity transform and optimize the values of the transformation from using recursive method to minimise the error of the similarity transform. The transform can be found as;

$$\lambda R l_i + t = r_i$$

$$R \ (rotation\ matrix) = \begin{bmatrix} cos\alpha & -sin\alpha \\ sin\alpha & cos\alpha \end{bmatrix} \quad (1)$$

$$t \ (translation\ matrix) = (t_x, t_y)$$

Equation (1) can be optimized by minimizing the sum of absolute error. The determined transform is then used remap the objects and correct the position as shown in the Fig.7. (a) and (b) shows how the objects have been remapped and agents path has been corrected.

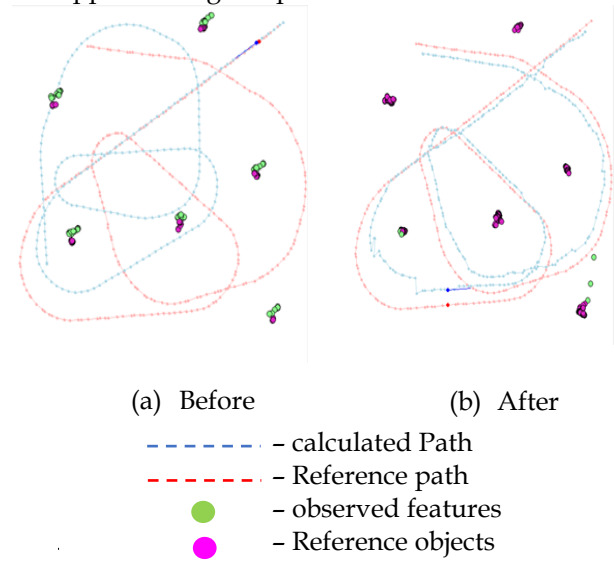

(a) Before  (b) After
– – – – – – calculated Path
– – – – – – Reference path
● – observed features
● – Reference objects

**Figure 8 – Applying Similarity Transform**

### 5.2 Particle Filter (Fast SLAM)

The major drawback of the similarity transform method is the difficulty to converge on optimised parameter value set in real-time. Hence, another method based on Bayesian filtering called Particle filter has been used. Particle filter SLAM [15] is a better solution, as the convergence rate is very high and with the probabilistic inferences to the estimation, it will address the uncertainty in the LRF too. Initially, it is hypothetically assumed that the agent will be anywhere with any orientation. Then Position estimate is computed using the density of the particles, representing the position with orientation. This is a two-step process,

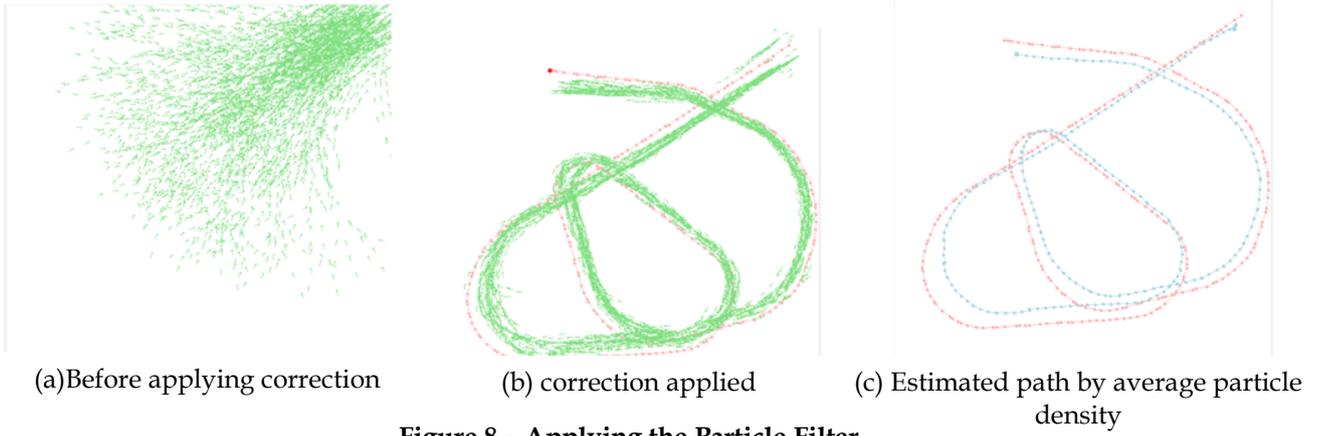

(a) Before applying correction  (b) correction applied  (c) Estimated path by average particle density

**Figure 8 – Applying the Particle Filter**

**A).** Prediction step- Particles are generated to represent the position, through iterative sampling of a normal distribution

$$\text{for } i = 1 \text{ to } n;$$
$$\text{sample } x^{(i)} \sim N(\mu, \sigma^2); \quad (2)$$

Then the particles are translated with the aid of the applied control signal and then resampled using resampling wheel technique.

It is a method that gives a higher chance to choose a particle from a position which has a higher weight (weighting is done in the correction step) rather than others.

$$\text{for } i = 1 \text{ to } n;$$
$$\text{sample } x^{(i)} \sim p(x_t \mid x_{t-1}^{[n]}, U_t); \quad (3)$$

**B).** Correction Step - This is done by weighting the particle by finding the match between obtained measurement and the position of the particle. This will reduce the outliers with the time and quickly converge to the actual path.

Drawn samples with probability $\alpha\ w_t^{[i]}$

$$\text{for } i = 1 \text{ to } n;$$
$$w_t^{[m]} \sim p(z_t \mid \bar{x}_t^{[m]}, U_t); \text{ where } z \text{ is LRF measurement}$$
$$p(z_t \mid \bar{x}_t^{[m]}) = \prod_i p(z_i \mid \bar{x}_t^{[m]}) \quad (4)$$
$$p(z_i \mid \bar{x}_t^{[m]}) = p(d - \hat{d}) * p(\alpha - \hat{\alpha})$$
$$= N(d - \hat{d}, 0, \sigma_d^2) * N(\alpha - \hat{\alpha}, 0, \sigma_\alpha^2)$$

$d - \hat{d}$ = error in the distance
$\alpha - \hat{\alpha}$ = error in orientation

After the calculations,

$$\text{position estimation} = 1/n * \sum_i x_i \quad (5)$$
$$\text{orientation estimation} = \tan^{-1}\left\{\frac{\sum_i \cos \vartheta_i}{\sum_i \sin \vartheta_i}\right\}$$

Through the above process, a particle filter SLAM can be implemented. A much accurate position estimation is rendered through that approach as seen from the Fig.8.

## 6. Conclusion

This work has been focussed on finding a simple and practically realizable solution to the engineering problem of constructing a map of an unknown environment. In the approach proposed, a low-cost laser range finder (LRF) has been selected as the sensor due its high accuracy along with simplicity when compared with traditional ultra-sonic sensors and complex vision systems. Methods have been developed for correcting the distortions and acquiring the data from the LRF. But any convenient sensor capable of producing an accurate distance data spectrum can be easily used to replace the LRF.

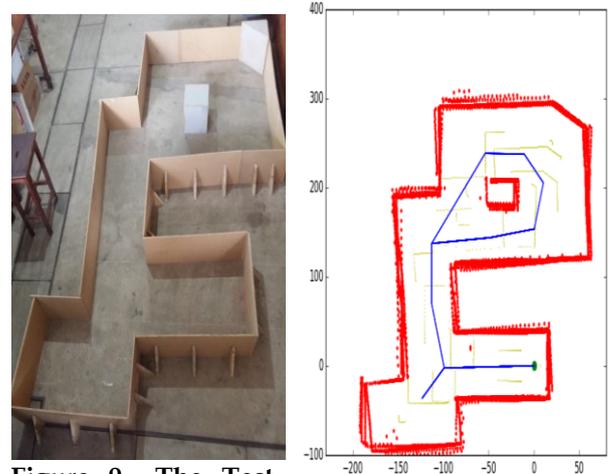

**Figure 9 – The Test Environment and Generated 2D Map.**

As the hardware platform, a 3WD Omni directional mobile platform has been chosen due to its convenient manoeuvrability into any direction intended. To guide it towards information gap in the data and to provide optimum path planning a separate navigation algorithm has been implemented. Key features like, 'Smallest gap first approach' and the 'Intelligent neighbour identification process' has allowed the mapping efficacy to be at optimum.

Due to the errors in the encoder based position measurement of the system, a parallelly running simultaneous localization and mapping (SLAM) algorithm has been developed. Similarity transform based approach and Monte-Carlo method (Particle filter) has been experimented and the Particle filter based Fast SLAM has been implemented on the system.

Altogether taken the combined work on above approaches has enabled to create a simple and rapidly realizable solution to the problem of environment mapping. Although specific hardware set-up is considered in this work, the developed algorithms are quickly adaptable to any other hardware combination. Also, as further improvements if a suitable sensor like a vision system is available, it would be convenient to extended the 2D map, generated through the proposed approach to a complete 3D map of the locale.

# Acknowledgement

This research work has been hosted and supported by the Department of Electrical and Electronic Engineering, Faculty of Engineering, University of Peradeniya, Sri Lanka.